\pdfoutput=1
\documentclass[11pt]{article}

\usepackage[preprint]{coling}

\usepackage{times}
\usepackage{latexsym}
\usepackage{multirow}

\usepackage[T1]{fontenc}
\usepackage[utf8]{inputenc}
\usepackage{microtype}
\usepackage{inconsolata}
\usepackage{graphicx}
\usepackage{amsmath}
\usepackage{booktabs}
\usepackage{algorithm}
\usepackage{algorithmic}

\usepackage{cleveref}
\crefname{section}{Sec.}{Secs.}
\Crefname{section}{Section}{Sections}
\Crefname{table}{Table}{Tables}
\crefname{table}{Tab.}{Tabs.}
\Crefname{figure}{Figure}{Figures}
\crefname{figure}{Fig.}{Figs.}
\Crefname{equation}{Equation}{Equations}
\crefname{equation}{Eq.}{Eqs.}

\title{Breaking the Stage Barrier: A Novel Single-Stage Approach to \\ Long Context Extension for Large Language Models}

\author{
 \textbf{Haoran Lian\textsuperscript{1}\thanks{These authors contributed equally to this work.}},
 \textbf{Junmin Chen\textsuperscript{2}\footnotemark[1]},
 \textbf{Wei Huang\textsuperscript{4}\footnotemark[1]},
 \textbf{Yizhe Xiong\textsuperscript{3}\footnotemark[1]},
 \textbf{Wenping Hu\textsuperscript{2}\footnotemark[1]}, \\
 \textbf{Guiguang Ding\textsuperscript{3$^{\dagger}$}}, 
 \textbf{Hui Chen\textsuperscript{3}},
 \textbf{Jianwei Niu \textsuperscript{1$^{\dagger}$}},
 \textbf{Zijia Lin\textsuperscript{2$^{\dagger}$}},
 \textbf{Fuzheng Zhang\textsuperscript{2}},
 \textbf{Di Zhang\textsuperscript{2}}
\\
\\
 \textsuperscript{1}Beihang University,
 \textsuperscript{2}Kuaishou Technology, 
 \textsuperscript{3}Tsinghua University,
 \\
 \textsuperscript{4}Beijing University of Posts and Telecommunications
\\
 \small{
   \textbf{Correspondence$^{\dagger}$:} \href{mailto:niujianwei@buaa.edu.cn}{niujianwei@buaa.edu.cn}, \href{mailto:dinggg@tsinghua.edu.cn}{dinggg@tsinghua.edu.cn}, \href{mailto:linzijia07@tsinghua.org.cn}{linzijia07@tsinghua.org.cn}
 }
}

\begin{document}
\maketitle
\begin{abstract}
Recently, Large language models (LLMs) have revolutionized Natural Language Processing (NLP). 
Pretrained LLMs, due to limited training context size, struggle with handling long token sequences, limiting their performance on various downstream tasks. Current solutions toward long context modeling often employ multi-stage continual pertaining, which progressively increases the effective context length through several continual pretraining stages. 
However, those approaches require extensive manual tuning and human expertise.
In this paper, we introduce a novel single-stage continual pretraining method, \textbf{H}ead-\textbf{A}daptive \textbf{R}otary \textbf{P}osition \textbf{E}ncoing (\textbf{HARPE}), to equip LLMs with long context modeling capabilities while simplifying the training process.
Our HARPE leverages different Rotary Position Encoding (RoPE) base frequency values across different attention heads and directly trains LLMs on the target context length.
Extensive experiments on 4 language modeling benchmarks, including the latest RULER benchmark, demonstrate that HARPE excels in understanding and integrating long-context tasks with single-stage training, matching and even outperforming existing multi-stage methods.
Our results highlight that HARPE successfully breaks the stage barrier for training LLMs with long context modeling capabilities.
% Long context modeling Large language models (LLMs), enabling them to capture complex relationships and nuances in text, which is crucial for tasks such as text summarization, question answering, and natural language generation. 
% Current state-of-the-art methods involve continual pre-training on longer sequences in a progressive extension strategy. 
% These multi-stage approaches require extensive hyperparameter tuning, requiring substantial computational resources and expertise. 
% In this paper, we propose a novel single-stage approach featuring Head-Adaptive Rotary Position Encoding (HARPE), which allows the model to adapt to varying context lengths in a single training stage. 
% Our method outperforms multi-stage continual training on multiple benchmarks, achieving state-of-the-art results at reduced training costs. 
% Notably, our approach demonstrates exceptional versatility, excelling in a wide range of tasks, including challenging "needle in a haystack" tasks, both single and multi-needle variants, as well as achieving lower perplexity on long documents and higher accuracy on short benchmark tasks, surpassing multi-stage extension approaches.
\end{abstract}

\section{Introduction}

% P1: Transformers, NLP and Long Context
In recent years, generative Large Language Models (LLMs) \cite{brown2020language,raffel2020exploring,touvron2023llama,fu2023kwaiyiimath,su2024maskmoe,xiong2024temporal,lian2024scaffold,lian2024lbpe} have dominated the field of Natural Language Processing, outperforming traditional task-specific methods on many tasks, like text summarization \cite{liu2019text,zaheer2020big,wu2021fastformer}, information extraction \cite{DBLP:conf/acl/WeiSZZGJ20,DBLP:conf/acl/WeiYJSZZLZLZ23,DBLP:journals/tkde/WeiSZJZLG23} and question answering \cite{brown2020language,raffel2020exploring}. 
% As a crucial approach for utilizing LLMs, in-context learning often requires LLMs to manage very long text sequences. 
In the process of utilizing LLMs for downstream tasks, it is often necessary for the LLM to handle long token sequences. 
For example, when conducting text summarization with an LLM, the input sequence may include an entire book \cite{zhang2024bench,karpinska2024one}, which contains millions of words.
%Popular open-source LLMs, such as the LLaMA series \cite{touvron2023llama, dubey2024llama}, are unable to digest long sequences, i.e., $>10$K tokens.
To equip LLMs with the capability to handle long texts, current methods typically continually pretrain LLMs on a larger context window \cite{xiong2023effective,peng2023yarn,fu2024data} compared to that in LLM pretraining. 
Given that Rotary Position Encoding (RoPE) \cite{su2024roformer} is the prevailing position encoding in most LLMs, among those methods, the mainstream approach is to increase the RoPE base frequency in the positional encoding during continual pretraining \cite{xiong2023effective}, as studies have demonstrated that a larger base frequency is the prerequisite for handling longer text sequences \cite{liu2023scaling,men2024base}.
%Through continual pretraining with a larger RoPE base frequency, recent studies have successfully extended the effective context length of LLMs to over 1 million tokens \cite{liu2024world}.

% P2: Motivation: prior works use multi-stage
%To achieve a big effective context size, existing works applied a multi-stage approach to extend the context length of LLMs, in which the effective context length is progressively increased through a series of continue pretraining steps. For example, Large World Model \cite{liu2024world} adopts a pipeline of increasing training context sizes 32K-128K-256K-512K-1M  to finally reach a 1M context window. The latest Llama 3.1 model series \cite{dubey2024llama} also adopt a six-stage pretraining pipeline to gradually increase the effective context length to 128k. Currently, multi-stage continual pretraining methods have dominated the community and have been the most common practice to equip LLMs with long context capabilities.

To achieve a large effective context size, existing works commonly employ a multi-stage approach, progressively increasing the context length through a series of continued pretraining steps. For instance, Large World Model \cite{liu2024world}, GLM4-Chat-1M \cite{glm2024}, MiniCPM-2.4B-128K \cite{2404.06395} and Llama 3.1 \cite{dubey2024llama} utilize multi-stage pipelines to reach context windows of 1M and 128k, respectively. This approach has become the dominant method in the community for equipping LLMs with long context capabilities.

%Although multi-stage approaches have demonstrated promising results on various benchmarks, our experiments reveal that substantial manual tuning and human expertise are demanded to achieve promising results on long context modeling benchmarks.
%For example, as shown in \cref{tab:multistage_choose}, when applying a three-stage ABF on the LLaMA model, simply training for the same amount of tokens in each stage significantly underperforms the pipeline in which training amount in each stage is carefully selected. On the long context modeling benchmark NiaH, carefully scheduling the training leads to a significant 13.5\% performance gain, indicating that the hyperparameters in multi-stage pipelines \cite{liu2024world} may lack generalizability across different LLM sizes and architectures.
%Furthermore, deploying a multi-stage training pipeline presents additional challenges due to the varying data and resource requirements at each stage. 
%Hence, a single-stage continual pretraining approach becomes critical for the practical development of long-context models.
%However, single-stage training also comes with its own challenges.
%For example, when an LLM is pretrained with a context window smaller than 10K, the first continual pretraining step typically expands the context window to 32K \cite{liu2024world}. 
%We also show in our experiments that training LLMs with a much longer context window, often paired with a much larger RoPE base value, leads to suboptimal outcomes, posing a dilemma for long-context continual pretraining.
Our single-stage experiments show that directly scaling a larger RoPE base in a single stage is less effective than using multi-stage approaches. This likely explains why most publicly available models employ multi-stage ABF (Adjusted Base Frequency) training. We hypothesize that direct scaling to the final training length without intermediate stages struggles to adapt to increased complexity, which is better managed through gradual, multi-stage adjustments.

Although multi-stage approaches have shown promising results, our experiments reveal that they require substantial manual tuning and human expertise to achieve good performance on long context modeling benchmarks. For instance, our results in \cref{tab:multistage_choose} demonstrate that a carefully scheduled three-stage pipeline outperforms a naive approach by 13.5\% on the NiaH benchmark, highlighting the limited generalizability of hyperparameters across different LLM sizes and architectures. Moreover, multi-stage training poses practical challenges due to varying data and resource requirements. This motivates the need for single-stage continual pretraining approaches. However, single-stage training also presents challenges, such as the risk of suboptimal outcomes when training with a much longer context window and larger RoPE base frequency, as shown in our experiments.

\begin{table}[t]
\centering
\resizebox{\columnwidth}{!}{
% \Large
\begin{tabular}{cccc}
\toprule
\multirow{2}{*}{\textbf{Metric}} & \multicolumn{2}{c}{\textbf{Three-Stage ABF on LLaMA}} \\
\cline{2-3}
& Uniform 2B Tokens & Carefully Selected \\
\midrule 
\textbf{NiaH} & 67.83 & \textbf{81.36} \\
\textbf{Benchmark} & 62.59 & \textbf{63.10} \\
\bottomrule
\end{tabular}}
\caption{
    Performance comparison of different continual pertaining pipelines when conducting three-stage ABF (Adjusted Base Frequency) \cite{xiong2023effective}. For both pipelines, we report the average score of the Needle-in-a-Haystack task (6 lengths, 8 tasks) and on Short-Context Benchmarks (5 tasks). For more details on the experiment settings, please refer to \cref{sec:exp_setup}.
}
\label{tab:multistage_choose}
\end{table}

In this paper, we introduce a novel \textbf{single-stage} approach, termed \textbf{H}ead-\textbf{A}daptive \textbf{R}otary \textbf{P}osition \textbf{E}ncoding (\textbf{HARPE}), designed to address the long-context problem. Our goal is to achieve an effective context length comparable to that of multi-stage methods.
Drawing inspiration from the finding that different attention heads can acquire distinct knowledge during training \cite{li2023functional}, we propose to distribute the training of different stages across multiple attention heads concurrently. Specifically, we leverage RoPE \cite{su2024roformer}  with varying base values to represent different effective context lengths, thereby simulating multiple training stages. By assigning different base values to different attention heads, we enable the LLMs to be trained in a single stage.

To determine the RoPE base values for each attention head, we employ a complementary approach, carefully selecting values that fill in the peaks and valleys of the sine and cosine waves in RoPE, thereby optimizing the experimental results.
In contrast to existing methods, our proposed HARPE offers a significant advantage in terms of simplicity and efficiency. By pretraining LLMs in a single stage, we substantially streamline the process of data preparation and pipeline adjustment, eliminating the need for multiple stages and associated complexities.

% Experimental & Significance (first to show that multi-stage can be shortened, reduce training cost, simplify the process)

We conduct a comprehensive evaluation of HARPE on 4 benchmarks, including the recently introduced RULER benchmark \cite{hsieh2024ruler}, to assess its effectiveness on both long-context and short-context tasks.
The experimental results demonstrate that HARPE consistently matches or surpasses the performance of its multi-stage counterparts across all evaluated benchmarks. Notably, on the challenging Needle-in-a-haystack task, HARPE achieves a significant improvement of 5.46\% over the multi-stage Adjused Base Frequency (ABF) \cite{xiong2023effective} approach, underscoring its exceptional capabilities in long-context modeling.

%We emphasize that our core contribution of  HARPE is the \textit{single-stage} continual pre-training strategy that successfully achieves a 128K effective context length. 128K有效长度没能达到，按照ruler，需要每个长度都比llama2 高才叫effective。

Unlike inference methods that employ multiple RoPE bases simultaneously to support long contexts \cite{zhang2024found,chen2024fortify}, our HARPE approach fundamentally alters the learning dynamics of LLMs during continual pretraining, enabling a straightforward and streamlined training pipeline.
In summary, our contributions are threefold:

\begin{itemize}

\item We introduce a novel single-stage continual pretraining approach, termed Head-Adaptive Rotary Position Encoding (HARPE), to address the long context problem in LLMs. By doing so, we significantly simplify the process of data preparation and pipeline adjustment.
\item To overcome the limitations of traditional multi-stage approaches, we propose a novel training strategy that distributes the training of different stages across multiple attention heads. We utilize different RoPE base values to represent distinct training stages and carefully select these values to complement the attention scores.
\item We conduct a comprehensive evaluation of HARPE on 4 long context benchmarks, including the recently introduced RULER benchmark. Our experimental results demonstrate that HARPE consistently yields comparable or even better performance than existing multi-stage methods across all benchmarks.%, achieving an effective context length of 128k.
\end{itemize}

\section{Related Works}
\textbf{Large Language Models (LLMs).} Language models are a type of statistical model that aims to maximize the likelihood token sequences \cite{touvron2023llama}. The Transformer architecture \cite{vaswani2017attention} marked a turning point in the evolution of language models, accelerating their development. Transformer-based models, like BERT \cite{devlin2018bert}, GPT-2 \cite{radford2019language}, and T5 \cite{raffel2020exploring}, have achieved groundbreaking results across numerous natural language processing tasks. More recently, the release of GPT-4 \cite{achiam2023gpt} has further pushed the boundaries of LLMs performance, showcasing exceptional capabilities. As these models continue to scale and evolve architecturally, they have become the driving force behind cutting-edge research in natural language processing, exhibiting notable adaptability and versatility across a wide range of applications \cite{liu2024deepseek,cai2024internlm2,team2024gemma，}. Consequently, LLMs have profoundly transformed human-computer interaction.

\noindent\textbf{Long Context Modeling.} Trained on relatively short context sequences (i.e., generally $<$10K tokens), open-source LLMs show dramatic performance drops on long context modeling \cite{touvron2023llama,bai2023qwen,liu2024deepseek}. Methods to improve the ability of LLMs to handle long context can be mainly divided into the following categories: attention mechanism optimizing, long-term memory caching, contexual processing, and positional encoding optimizing. Attention mechanism optimizing methods \cite{beltagy2020longformer,ma2021luna,dao2022flashattention} reduce the computational and memory bottlenecks of the Transformer, thereby enabling the model to process longer text sequences. Long-term memory caching methods \cite{wang2024augmenting,bulatov2022recurrent,martins2020sparse,dai2019transformer} utilize internal or external memory caches to fetch information in long context. Contextual processing methods \cite{ding2020cogltx,izacard2020leveraging} process long context inputs by calling the model multiple times to process different parts of the long text sequence.

\begin{figure*}[t]
\centering
\includegraphics[width=0.95\textwidth]{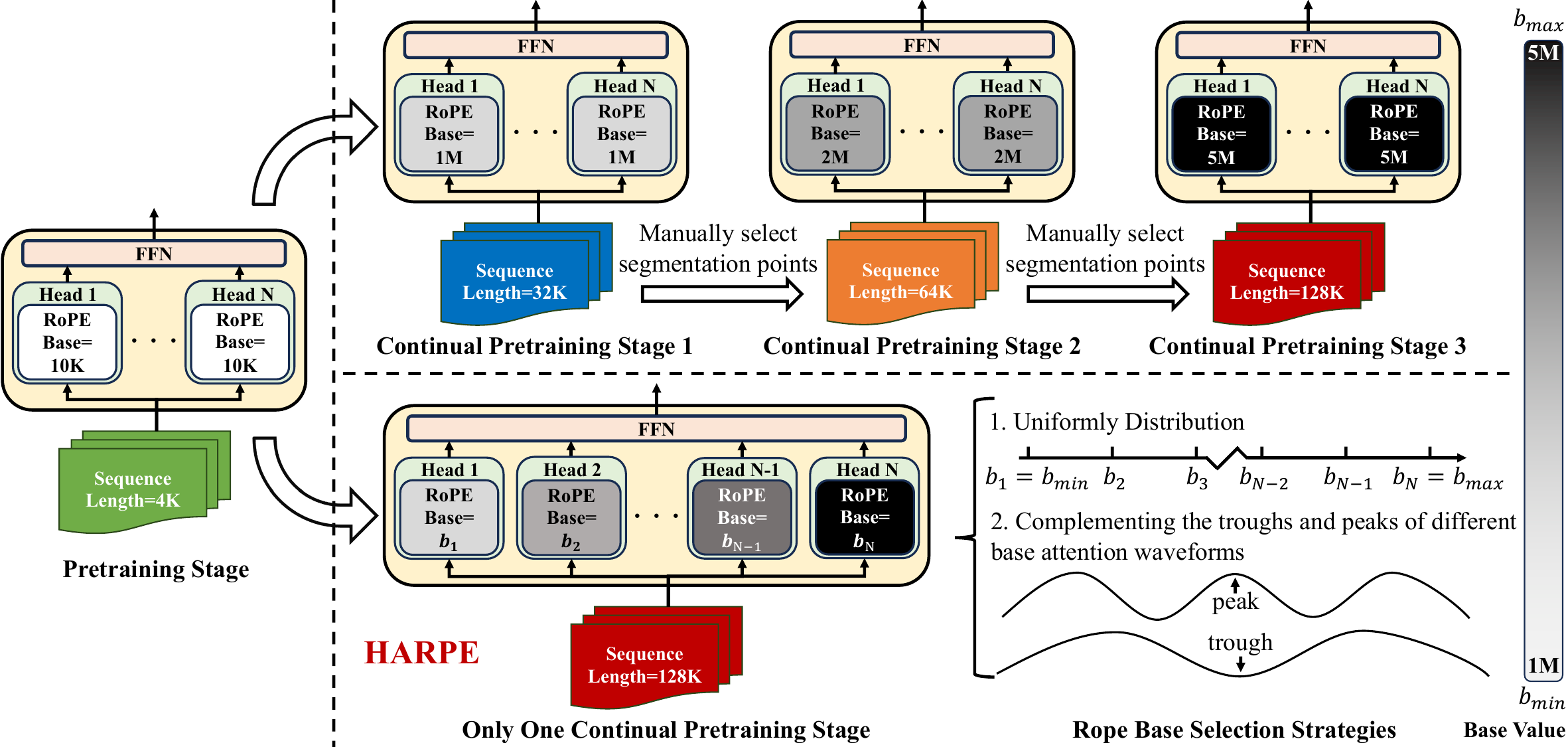}
\caption{Illustration of the multi-stage  and our proposed single-stage (HARPE) continual pretraining pipeline.}
\label{pipeline}
\end{figure*}   

Apart from those methods, the most common approach is to improve the RoPE \cite{su2024roformer} while conducting continual pretraining with a longer context window. Specifically, Position Interpolation (PI) \cite{chen2023extending} reduces the input position index to match the original context window size. ABF \cite{xiong2023effective} adjusts the RoPE base (i.e., $\theta$) to scale the low-frequency part more significantly, thereby dispersing the interpolation pressure to multiple dimensions. NTK-by-parts interpolation \cite{NTKAwarepart} interpolates RoPE bases according to the wavelength of different dimensions in RoPE relative to the context size: high-frequency dimensions are not interpolated, low-frequency dimensions are fully interpolated, and intermediate frequency dimensions are partially interpolated using a ramp function. YARN \cite{peng2023yarn} combines the NTK-by-parts interpolation with the attention scaling technique to achieve an even longer effective context length. Self-Extend \cite{jin2024llm} constructs a two-layer attention mechanism, consisting of group attention and neighbor attention, to successfully expand the context window without additional training. 
Studies have also explored the construction \cite{chen2024long} and training strategies \cite{bai2024longalign} of long context data.

While methods based on improving positional encoding have achieved promising experimental results \cite{liu2024world,he2024never,zhang2024explore}, they typically rely on complicated multi-stage training pipelines to gradually increase the effective context length(e.g., $8k \rightarrow 16k \rightarrow 32k \ldots \rightarrow 128k$). In contrast, our proposed HARPE offers a simplified \textit{single-stage} continual pretraining approach. %, which increases the effective context length of LLMs to \textbf{128K} by dynamically adjusting different RoPE bases on different attention heads. 
Experimental results demonstrate that our HARPE achieves comparable performance to existing multi-stage methods.

\section{Head-Adaptive Rotary Position Encoding based Approach}
%In this section, we first formulate Rotary Position Encoding (RoPE) in \cref{Preliminaries}. And then we discuss the proposed HARPE in \cref{HARPE} with its muti-head RoPE base mechanism and RoPE base selection strategies. 

In this section, we first revist the formulation of Rotary Position Encoding (RoPE) in \cref{Preliminaries}. We then present the proposed HARPE in \cref{HARPE}, detailing its multi-head RoPE base mechanism and base selection strategies.

\subsection{Preliminaries}
\label{Preliminaries}
RoPE \cite{su2024roformer} is a widely adopted technique for position encoding in LLMs built on the Transformer architecture \cite{glm2024chatglm,liu2024deepseek,yang2023baichuan,dubey2024llama}. The primary objective of RoPE is to encode positional information in a way that the inner product of the query and key embeddings inherently captures the relative position information, which can be formally expressed as:
\begin{equation}
f(q_m,m)^Tf(k_n,n)=g(q_m,k_n,m-n)
\end{equation}
Here, $f$ represents the positional encoding function applied to the query embeddings $q_m$ at position $m$ and key embeddings $k_n$ at position $n$. To satisfy this condition, the function $f$ is defined as a $d$-dimensional rotation matrix, denoted as
$\boldsymbol{R}_{\Theta,m}^d$:

\begin{equation}
f(x_{\{q,k\}},m)=\boldsymbol{R}_{\Theta,m}^dx_{\{q,k\}}
\end{equation}
where 
\begin{equation}
\boldsymbol{R}_{\Theta,m}^d = \operatorname{diag}\left(\boldsymbol{R}(m\theta_{1}), \ldots, \boldsymbol{R}(m\theta_{d/2})\right)
\end{equation}
\begin{equation}
\boldsymbol{R}(m\theta_{i}) = \begin{pmatrix} \cos m\theta_{i} & -\sin m\theta_{i} \\ \sin m\theta_{i} & \cos m\theta_{i} \end{pmatrix}
\end{equation}
\begin{equation}
\Theta=\{\theta_i=b^{-2(i-1)/d},i\in[1,2,\dots,d/2]\} \label{eq:theta}
\end{equation}

These formulas indicate that the rotation angle $\Theta$ can be adjusted by modifying the base frequency $b$. Specifically, increasing $b$ (i.e., decreasing $\Theta$) mitigates the severe decaying effect of RoPE on attention scores for distant tokens, thereby enabling LLMs to process longer input sequences \cite{xiong2023effective}.
%Indeed, existing approaches to extending the LLM context length have relied on gradually increasing both the base $b$ and input sequence length $l$ in stages during continual pretraining \cite{dubey2024llama,liu2024world}. However, this multi-stage continual pretraining method requires labor-intensive manual selection of stage segmentation points. To address this limitation, we propose HARPE, a novel approach that distributes multiple bases across all attention heads in a single layer, effectively integrating multi-stage training into a single-stage training paradigm.

%Such formulas mean that we can change $\Theta$ by changing the base frequency $b$. And increasing $b$ (i.e., decreasing the rotation angle $\Theta$) which reduces the heavy decaying effect of RoPE on the attention scores for distant tokens can help LLMs process longer input sequences \cite{xiong2023effective}. 

%In fact, the existing works of extending the LLM context length are achieved by gradually increasing both the base $b$ and input sequences length $l$ in stages during continual pretraining \cite{dubey2024llama,liu2024world}. 
%However, such muti-stages continual pre-training method requires laboriously manual selection of stage segmentation points. To address that issue, we propose HARPE, which distributes multiple bases on all attention heads in a single layer to integrate multi-stage training into single-stage training.

\subsection{HARPE}
\label{HARPE}
% \subsubsection{Muti-Head RoPE Base}
% \label{Muti-Head-RoPE-Base}
%As shown in \cref{pipeline}, to utilize the different abilities of each attention head, HARPE pre-determines a base set $B$ with $|B|$ is equal to the number of attention heads $N$ in a single layer. Then HARPE assigns a base $b_h$ in $B$ to RoPE in each attention head $h$.
As illustrated in \cref{pipeline}, HARPE leverages the diverse capabilities of each attention head by pre-defining a base set $B$, where the cardinality of $B$ is equal to the number of attention heads. Specifically, HARPE assigns a unique base $b_h$ from $B$ to the RoPE in each attention head $h$.
For head $h$,  $\Theta$ in Eq. \ref{eq:theta} is defined as:
\begin{equation}
\Theta_{h}=\{\theta_{h,i}=b_{h}^{-2(i-1)/d},i\in[1,\dots,d/2]\}
\end{equation}
then
\begin{equation}
\boldsymbol{R}_{\Theta_{h},m}^d = \operatorname{diag}\left(\boldsymbol{R}(m\theta_{h,i})\right)
\end{equation}

\begin{equation}
\boldsymbol{R}(m\theta_{h,i}) = \begin{pmatrix} \cos m\theta_{h,i} & -\sin m\theta_{h,i} \\ \sin m\theta_{h,i} & \cos m\theta_{h,i} \end{pmatrix}
\end{equation}

% \subsubsection{RoPE Base Selection Strategy}
% \label{RoPE-Base-Selection-Strategy}
%As for the determination of $B$, we set up two different RoPE base selection strategies. The first strategy is to evenly distribute the bases within the range of a maximum base $b_{max}$ and minimum base $b_{min}$:
To determine the base set $B$, we establish two distinct RoPE base selection strategies. 

The first strategy involves uniformly distributing the bases $B_{uniform}$ within a predefined range, bounded by a maximum base $b_{max}$ and a minimum base $b_{min}$.
\begin{equation}
B_{uniform}=\{b_{h}=b_{min} + h \times \frac{b_{max}-b_{min}}{N - 1}\}
\end{equation}
where $h = 0, 1, \dots, N-1$.

% 讲一下motivation
The second strategy adopts the search method proposed by \cite{chen2024fortify}, which seeks to ensure that the attention waveform valleys of any given base overlap with peaks from different bases, and vice versa. To achieve this, a candidate base set $B_c$ is initially generated by discretizing the range between $b_{max}$ and $b_{min}$ with a relatively small stride $s$.

%The second strategy is the search method of \cite{chen2024fortify} which aims to ensure that the attention waveform troughs of any given base overlap with peaks from different bases, and vice versa. Firstly, a large base candidate set $B_c$ is generated within the range of $b_{max}$ and $b_{min}$ with a relatively small stride $s$:
\begin{equation}\label{eq:bc}
B_c=\{b_{\min}+i\times s, i\in[1,\frac{b_{\max}-b_{\min}}s]\}
\end{equation}
Subsequently, the final searched base set $B_s$ is determined by iteratively complementing the valleys and peaks of attention waveforms of different bases within $B_c$, as shown in Algorithm \ref{The searching algorithm of B_s}.

\begin{algorithm}[t]
\caption{The searching algorithm of $B_s$}
\label{The searching algorithm of B_s}
\begin{algorithmic}[1]
\REQUIRE A candidate base set $B_c$
\STATE Define a function $f_p(b) \mapsto$ peak positions in attention waveforms corresponding to base $b$
\STATE Define a function $f_v(b) \mapsto$ valley positions in attention waveforms corresponding to base $b$
% Calculate the peak positions $P_c$ and trough positions $T_c$ in attention waveforms corresponding to bases in $B_c$
\STATE Initialize the searched base set $B_s \leftarrow \{b_{min}\}$
\STATE $P_s \leftarrow f_p(b_{min});V_s \leftarrow f_v(b_{min})$
\WHILE{$|B_s| < N$}
    \FOR{$b_j$ in $B_c$}
        \STATE $P_j \leftarrow f_p(b_j);V_j \leftarrow f_v(b_j)$
        \STATE $d_j^+\leftarrow \sum_{\begin{array}{c}p_{j,i}\in P_j\\v_{s,i}\in V_s\end{array}}|p_{j,i}-v_{s,i}|$
        \STATE $d_j^-\leftarrow \sum_{\begin{array}{c}v_{j,i}\in V_j\\p_{s,i}\in P_s\end{array}}|v_{j,i}-p_{s,i}|$
        \STATE $d_j \leftarrow d_j^+ + d_j^-$
    \ENDFOR
    \STATE $B_s \leftarrow B_s \cup \{b'_j$ with the minimum $d_j\}$
    \STATE $P_s \leftarrow P_s \cup f_p(b'_j);V_s \leftarrow V_s \cup f_v(b'_j)$
\ENDWHILE
\RETURN $B_s$
\end{algorithmic}
\end{algorithm}

% \begin{table}[t]
% \centering
% \resizebox{\columnwidth}{!}{
% \begin{tabular}{cc|cc|cc|cc}
% \toprule
% \textbf{Head} & \textbf{Base} & \textbf{Head} & \textbf{Base} & \textbf{Head} & \textbf{Base} & \textbf{Head} & \textbf{Base} \\
% \midrule
% 1 & $1.00 \times 10^6$ & 9 & $2.50 \times 10^6$ & 17 & $3.01 \times 10^6$ & 25 & $3.61 \times 10^6$ \\
% 2 & $1.15 \times 10^6$ & 10 & $2.65 \times 10^6$ & 18 & $3.04 \times 10^6$ & 26 & $3.88 \times 10^6$ \\
% 3 & $1.30 \times 10^6$ & 11 & $2.68 \times 10^6$ & 19 & $3.10 \times 10^6$ & 27 & $4.09 \times 10^6$ \\
% 4 & $1.45 \times 10^6$ & 12 & $2.71 \times 10^6$ & 20 & $3.13 \times 10^6$ & 28 & $4.15 \times 10^6$ \\
% 5 & $2.17 \times 10^6$ & 13 & $2.74 \times 10^6$ & 21 & $3.16 \times 10^6$ & 29 & $4.39 \times 10^6$ \\
% 6 & $2.20 \times 10^6$ & 14 & $2.80 \times 10^6$ & 22 & $3.22 \times 10^6$ & 30 & $4.45 \times 10^6$ \\
% 7 & $2.23 \times 10^6$ & 15 & $2.83 \times 10^6$ & 23 & $3.43 \times 10^6$ & 31 & $4.51 \times 10^6$ \\
% 8 & $2.47 \times 10^6$ & 16 & $2.92 \times 10^6$ & 24 & $3.46 \times 10^6$ & 32 & $4.54 \times 10^6$ \\
% \bottomrule
% \end{tabular}
% }
% \caption{
% % RoPE bases corresponding to each attention head in HARPE.
% RoPE base frequency settings for each head in HARPE, with stride of 30k.
% }
% \label{tab:rope_base}
% \end{table}

\begin{table}[t]
\centering
\resizebox{\columnwidth}{!}{
\begin{tabular}{cc|cc|cc|cc}
\toprule
\textbf{Head} & \textbf{Base} & \textbf{Head} & \textbf{Base} & \textbf{Head} & \textbf{Base} & \textbf{Head} & \textbf{Base} \\
\midrule
1 & 1.00 & 9 & 2.50 & 17 & 3.01 & 25 & 3.61 \\
2 & 1.15 & 10 & 2.65 & 18 & 3.04 & 26 & 3.88 \\
3 & 1.30 & 11 & 2.68 & 19 & 3.10 & 27 & 4.09 \\
4 & 1.45 & 12 & 2.71 & 20 & 3.13 & 28 & 4.15 \\
5 & 2.17 & 13 & 2.74 & 21 & 3.16 & 29 & 4.39 \\
6 & 2.20 & 14 & 2.80 & 22 & 3.22 & 30 & 4.45 \\
7 & 2.23 & 15 & 2.83 & 23 & 3.43 & 31 & 4.51 \\
8 & 2.47 & 16 & 2.92 & 24 & 3.46 & 32 & 4.54 \\
\bottomrule
\end{tabular}
}
\caption{
% RoPE base frequency settings for each head in HARPE, with stride of 30k.
RoPE base frequency settings for each head in HARPE, with each base value expressed in millions ($\times 10^6$), and stride of $30k$.
}
\label{tab:rope_base}
\end{table}

% Initially:
% \begin{equation}
% B_s=\{b_{min}\}
% \end{equation}
% repeat:
% \begin{equation}
% b_s \in B_c \text{ with maximum complementarity}
% \end{equation}
% \begin{equation}
% B_s=B_s\cup\{b_{s}\}
% \end{equation}
% until:
% \begin{equation}
% |B_s| = N
% \end{equation}

\section{Experimental Setup}
\label{sec:exp_setup}

We select LLama2-7B-Base \cite{touvron2023llama2} as our base model, which is configured with a RoPE base frequency of $10k$ and a context length of $4k$.

\subsection{Baseline Systems}\label{sub:baseline_systems}

We compare HARPE with 4 continual pretraining methods and one training-free method.

\textbf{PI} \cite{chen2023extending} employs a linear down-scaling of the input position indices to match the original context window size, thereby avoiding extrapolation beyond the trained context length, which can lead to catastrophically high attention scores that compromise the self-attention mechanism. The interpolation scale is set to $32=128k/4k$.

\textbf{ABF Single-Stage} \cite{xiong2023effective} implements a minimal yet necessary modification to the RoPE positional encoding for long-context modeling: increasing the hyperparameter "base frequency" $b$ to $5m$ (i.e., decreasing the rotation angle), which mitigates the decaying effect of RoPE for distant tokens. Concurrently, the input sequence length is increased to $128k$.

\textbf{ABF Multi-Stage} also increases the base and input sequence length, but with a key difference: it does so in a gradual, multi-stage manner. Specifically, we divide the process into three stages: $(1) b=1m; l=32k$, $(2) b=2m; l=64k$, and $(3) b=5m; l=128k$.

\textbf{YaRN} \cite{peng2023yarn} utilizes the RoPE formula \cref{eq:theta} to distinguish between high-frequency and low-frequency positional components. It adjusts the base within a 64-dimensional space according to these frequency components, applying a scale factor of 32. 

\textbf{Self-Extend} \cite{jin2024llm} is a training-free method. We apply it with $window\_size=1024$ and $group\_size=32$.

\begin{table}[t]
  \centering
  \resizebox{\columnwidth}{!}{
  \begin{tabular}{ccc}
    \toprule
    \centering
    \textbf{Method}       & \textbf{Proof-pile}& \textbf{GovReport} \\
    \midrule
    Llama2-7B-Base & 4336.96 & 7289.38  \\
    \midrule
    PI & 20.73 & 11.47  \\
    ABF Single-Stage & 3.06 & 3.58 \\
    ABF Multi-Stage & 3.03 & 3.57  \\
    YaRN & 4.53 & 4.52  \\
    \midrule
    Self-Extend* & 5.45 & 3.76  \\
    \midrule
    HARPE(ours) & \textbf{3.02} & \textbf{3.54} \\
    
    \bottomrule
  \end{tabular}
  }
  \caption{
    Sliding window perplexity (S = 256) for \textbf{Proof-pile} and \textbf{GovReport} documents. Asterisks (*) denote the training-free method. Lower perplexity values indicate better model performance.
    }
  \label{tab:GovReport}
\end{table}

\begin{figure*}[h]
\centering
\rotatebox{270}{\includegraphics[trim=80 5 110 87, clip, width=0.55\textwidth]{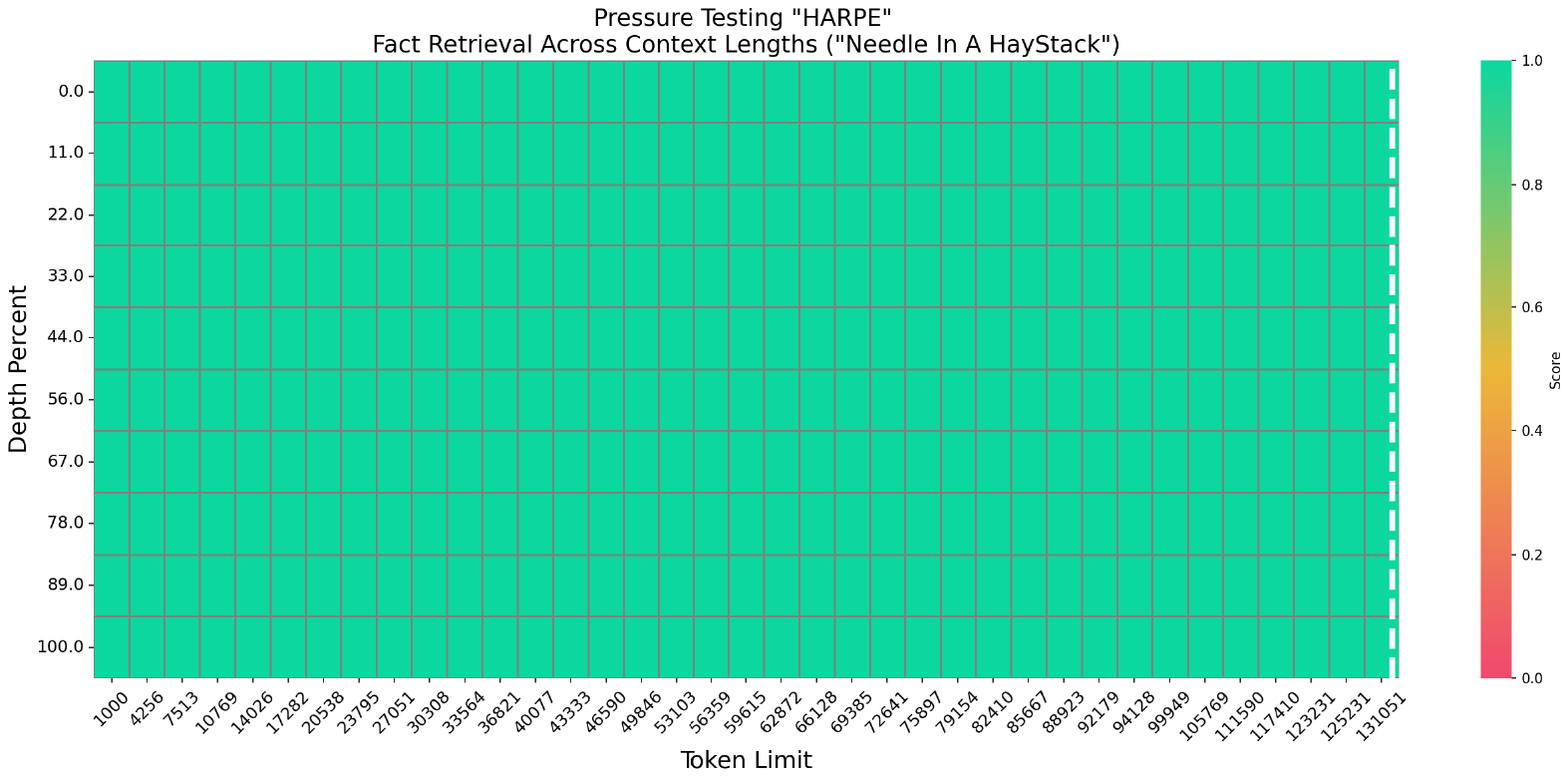}}
\caption{\textbf{Traditional Single-Key Needle-in-a-Haystack}: the x-axis represents the number of tokens in the test sample, ranging up to 128k tokens with finer granularity. The y-axis shows the depth of the needle's position within the current test sample. }
\label{passkey_retrieval}
\end{figure*}   
% \begin{figure}[t]
% \centering
% \includegraphics[width=0.55\textwidth, trim=0cm 0cm 0cm 0cm, clip]{passkey_retrieval.pdf}
% \caption{\textbf{Single-Key Needle-in-a-Haystack}: the x-axis represents the number of tokens in the test sample, ranging from 1000 to 128k tokens with intervals of 5k tokens. The y-axis indicates the depth of the needle's position within the current test sample. }
% \label{passkey_retrieval}
% \end{figure}  

\begin{table*}[t]
  \centering
  % \resizebox{\textwidth}{!}{
  \begin{tabular}{cccccccl}
    \toprule
    \centering
    \textbf{Method} & \textbf{4k} & \textbf{8k} & \textbf{16k} & \textbf{32k} & \textbf{64k} & \textbf{128k} & \textbf{Avg.} \\
    \midrule
    Llama2-7B-Base & 90.90 & - & - & - & - & - & ~~~- \\
    \midrule
    PI  & 77.56 & 26.59 & 16.50 & 0.00 & 0.00 & 0.00 & 20.11 \\
    ABF Single-Stage & 92.44 & 88.78 & 84.16 & 78.03 & 70.81 & 62.72 & $\text{79.49}_{\text{(3rd)}}$ \\
    ABF Multi-Stage  & 95.19 & 91.72 & 87.53 & 78.84 & 72.78 & 62.13 & $\text{81.36}_{\text{(2nd)}}$ \\
    YaRN  & 83.88 & 73.66 & 64.84 & 46.53 & 12.69 & 0.00 & 46.93 \\
    \midrule
    Self-Extend*  & 76.47 & 66.25 & 58.84 & 52.16 & 1.38 & 0.00 & 42.52 \\
    \midrule
    HARPE(ours)  & \textbf{97.03} & \textbf{96.88} & \textbf{93.72} & \textbf{86.66} & \textbf{79.41} & \textbf{67.19} & $\textbf{86.82}_{\text{(1st)}}$ \\
    %_{(1st)} \textbf{$86.82_{(1st)}$} $79.49_{(3rd)}$ 
    \bottomrule
  \end{tabular}
  % }
  \caption{
    \textbf{Upgraded Needle-in-a-Haystack Tests}: Average scores for 8 NiaH tasks at various lengths. Asterisks (*) denote training-free methods.
    }
  \label{tab:niah_methods_compare}
\end{table*}

\subsection{HARPE Base Setting}
We adopt the second strategy (i.e.,the peak-valley search method) mentioned in Sec. \ref{HARPE}. We set $b_{min}=1m;b_{max}=5m;s=30k$. The final bases are shown in Tab. \ref{tab:rope_base}. And we will discuss other base settings in Sec. \ref{subsec:52_study}.

\subsection{Evaluation Metric}\label{subsec:evaluation_metric}

\textbf{Perplexity (PPL)} is evaluated on the Proof-pile \cite{proof-pile} and GovReport \cite{huang2021efficient} datasets. % using the formula $ppl=e^{loss}$, following the settings outlined in YaRN \cite{peng2023yarn}. 
Following the setup in Yarn, for the Proof-pile dataset, we selected samples with a minimum of 128k tokens and measured perplexity for token lengths ranging from 2k to 128k in increments of 2k, averaging the scores for each length. For the GovReport dataset, we reported the average PPL scores for samples with a context window of 32k tokens. Evaluations are conducted using the sliding window method proposed by Press \cite{press2021train}, with a window size of 256 tokens. %The results are summarized in \cref{tab:GovReport}.

\textbf{Needle-in-a-Haystack} is a task that assesses a model's ability to accurately locate and recite a specific sentence, referred to as the "needle", within a lengthy document, known as the "haystack". To provide a more comprehensive evaluation of a model's long-context capabilities, we extend this method, inspired by RULER \cite{hsieh2024ruler}, to include multi-key, multi-value and multi-query scenarios, as well as diverse types of needles and background documents in each scenario. A multi-key task involves multiple keys, similar to 'the needle', in the background, where the model must find the target needle among the distractions. In a multi-value task, multiple needles are inserted in haystack, and the model earns one point for each correct needle found.

\textbf{Short-Context Benchmarks} assess whether short-context capabilities are preserved during long-context training. We include five widely used short-context evaluation datasets: 5-shot MMLU \cite{2009.03300}, 10-shot Hellaswag \cite{zellers2019hellaswag}, 25-shot ARC-Challenge \cite{1803.05457}, 0-shot PiQA \cite{1911.11641}, and 5-shot TriviaQA \cite{1705.03551}.

\subsection{Training Configuration}\label{subsec:training_onfiguration}
For continual pretraining, we follow the configurations outlined in \cite{2402.10171}, utilizing the upsampling dataset from \cite{yaofu2023slimpajama}. We employ the Llama2-7B-Base model as the pre-trained backbone, with a learning rate of $2 \text{e}^{-5}$ and AdamW optimizer settings of $\beta_{1} = 0.9$ and $\beta_{2} = 0.95$. All models were continually pre-trained with 6B tokens using these consistent settings. 

\begin{table}[t]
  \centering
  \resizebox{\columnwidth}{!}{
    \begin{tabular}{cccc}
    \toprule
    \multirow{2}*{\textbf{Method}} & \textbf{ABF} & \textbf{ABF} & \textbf{HARPE} \\
    ~ & \textbf{Single-Stage} & \textbf{Multi-Stage} & \textbf{(Ours)} \\
    \midrule
    \textbf{MMLU} & 40.87 & 41.10 & 40.74 \\
    \textbf{Hellaswag} & 77.33 & 77.83 & 77.99 \\
    \textbf{ARC-c} & 52.82 & 52.65 & 52.73 \\
    \textbf{PIQA} & 78.56 & 78.56 & 78.56 \\
    \textbf{TriviaQA} & 62.39 & 63.29 & 63.72 \\
    \midrule
    \textbf{Avg.} & 62.39 & 62.69 & 62.75 \\
    \bottomrule
  \end{tabular}
  }
  \caption{
   \textbf{Short-Context Benchmark Results}: Evaluation Results of the Top 3 Long-Context-Performance Models on 5 Short-Context Datasets.
  }
    \label{tab:benchmark_methods_compare}
\end{table}

\begin{table*}[htbp]
  \centering
  \resizebox{\textwidth}{!}{
  \begin{tabular}{ccccccccl}
    \toprule
    \textbf{Method}           & \textbf{Model}       & \textbf{4k} & \textbf{8k} & \textbf{16k} & \textbf{32k} & \textbf{64k} & \textbf{128k} & \textbf{Avg.} \\
    \midrule
    RoPE & Llama2-7B-Base & 90.9 & - & - & - & - & - & ~~~- \\
    \midrule
    \multirow{5}*{peak\&valley} & $stride=10k$ & 96.38 & 96.38 & 88.97 & 84.09 & 76.00 & 65.22 & 84.51 \\
    complementarity & $stride=20k$ & 96.97 & 96.78 & 91.69 & 85.97 & 74.44 & 67.59 & $\text{85.57}_{\text{(2nd)}}$ \\
    ~ & $stride=30k$ & 97.03 & 96.88 & 93.72 & 86.66 & \text{79.41} & 67.19 & $\text{86.82}_{\text{(1st)}}$ \\
    ~ & $stride=40k$ & 96.84 & 96.09 & 91.13 & 83.13 & 75.94 & 65.00 & 84.69 \\
    ~ & $stride=50k$ & 92.75 & 92.4 & 87.31 & 83.97 & 75.31 & 70.34 & 83.68\\
    \midrule
    \multirow{2}*{same stride} & ascending order & 96.53 & 95.13 & 89.22 & 84.72 & 74.31 & 65.88 & 84.30 \\
   ~ & descending order & 96.50 & 95.22 & 91.91 & 87.16 & 76.31 & 64.69 & $\text{85.30}_{\text{(3rd)}}$ \\
    \bottomrule
  \end{tabular}
  }
  \caption{\textbf{Upgraded Needle-in-a-Haystack Results of HARPE}: Comparison of different base selection        schemes in HARPE models. 
    }
  \label{tab:niah_ablation}
\end{table*}

% \begin{table*}[htbp]
%   \centering
%   \begin{tabular}{ccccccccl}
%     \toprule
%     \textbf{Method}           & \textbf{Model}       & \textbf{4k} & \textbf{8k} & \textbf{16k} & \textbf{32k} & \textbf{64k} & \textbf{128k} & \textbf{Avg.} \\
%     \midrule
%     RoPE & Llama2-7B-Base & 90.9 & - & - & - & - & - & ~~~- \\
%     \midrule
%     \multirow{5}*{peaks\&troughs} & $stride=10k$ & 96.38 & 96.38 & 88.97 & 84.09 & 76.00 & 65.22 & 84.51 \\
%     complementarity & $stride=20k$ & 96.97 & 96.78 & 91.69 & 85.97 & 74.44 & 67.59 & $\text{85.57}_{\text{(2nd)}}$ \\
%     ~ & $stride=30k$ & 97.03 & 96.88 & 93.72 & 86.66 & 79.41 & 67.19 & $\text{86.82}_{\text{(1st)}}$ \\
%     ~ & $stride=40k$ & 96.84 & 96.09 & 91.13 & 83.13 & 75.94 & 65.00 & 84.69 \\
%     ~ & $stride=50k$ & 92.75 & 92.4 & 87.31 & 83.97 & 75.31 & 70.34 & 83.68\\
%     \midrule
%     \multirow{2}*{same stride} & ascending order & 96.53 & 95.13 & 89.22 & 84.72 & 74.31 & 65.88 & 84.30 \\
%    ~ & descending order & 96.50 & 95.22 & 91.91 & 87.16 & 76.31 & 64.69 & $\text{85.30}_{\text{(3rd)}}$ \\
%     \bottomrule
%   \end{tabular}
%   \caption{
%     Comparison of different base selection schemes in HARPE models. 
%   }
%   \label{tab:niah_ablation}
% \end{table*}

\begin{table*}[h!]
  \centering
  \resizebox{\textwidth}{!}{
  \begin{tabular}{ccccccccl}
    \toprule
    \textbf{Model}   & \textbf{Size}  & \textbf{4k} & \textbf{8k} & \textbf{16k} & \textbf{32k} & \textbf{64k} & \textbf{128k} & \textbf{Avg.} \\
    \midrule
    Jamba \cite{ai21_jamba_2024} & 52B & 81.20 & 75.4 & 68.8 & 65.3 & 61.00 & 51.4 & 67.18 \\
    Mixtral \cite{mistral_platform_2023} & 7B & 91.60 & 89.80 & 86.30 & 77.20 & 52.30 & 8.00 & 67.50 \\
    % Mixtral \cite{2310.06839} & 8x7B & 91.80 & 91.00 & 89.50 & 85.80 & 66.90 & 29.00 & $\text{75.70}_{\text{(1st)}}$ \\
    \midrule
    Llama2-7B-Base & 7B & 79.4 & - & - & - & - & - & ~~~- \\
    Together \cite{togetherai_2023} & 7B & 84.6 & 78.7 & 68.3 & 57.9 & 0.0 & 0.0 & 48.25 \\
    Yarn \cite{peng2023yarn} & 7B & 77.30 & 67.50 & 59.00 & 47.30 & 38.60 & 13.90 & 50.60 \\
    LongLoRA \cite{chen2024longlora} & 7B & 81.90 & 80.4 & 75.6 & 65.1 & 60.80 & 0.0 & 60.63 \\
    LWM \cite{liu2024world} & 7B & 77.50 & 74.00 & 69.60 & 64.60 & 61.30 & 59.00 & $\text{67.67}_{\text{(3rd)}}$ \\
    llama-2-7b-80k \cite{yaofu_llama2_2023} & 7B & 87.95 & 80.68 & 72.70 & 63.47 & 54.62 & 47.65 & $\text{67.85}_{\text{(2nd)}}$ \\
    HARPE (ours) & 7B & 88.48 & 83.44 & 74.87 & 68.10 & 55.64 & 51.88 & $\text{70.40}_{\text{(1st)}}$ \\
    \bottomrule
  \end{tabular}
  }
  \caption{
    \textbf{RULER Benchmark Results}: Comparison of HARPE and Open-Source Base Models Across All Lengths for 13 RULER Tasks.
  }
  \label{tab:opensource_compare}
\end{table*}

\section{Experimental Results}
    \subsection{HARPE vs. Baseline Systems}\label{subsec:52}
    We utilize HARPE to conduct a comparative evaluation with the five long-context methods outlined in \cref{sub:baseline_systems}, employing three evaluation metrics as detailed in \cref{subsec:evaluation_metric}.
    
        %From the Proof-pile test set, we randomly selected 10 samples, ensuring each sample contained at least 128k tokens. We then evaluated the Perplexity scores for these samples at token lengths ranging from 2k to 128k, with a step size of 2k, and calculated the mean of all results. In the GovReport test set, we reported the average Perplexity scores for 50 samples with a context window of 32k tokens.The results are summarized in \cref{tab:GovReport}. It is noteworthy that all evaluations employed the sliding window method proposed by Press \cite{press2021train}, with a window size of 256. %

        % XYZ: 加到前面 First, we utilize the PPL metric to evaluate the long context modeling capability of HARPE.
        First, to evaluate the long context modeling capability of HARPE, we evaluate HARPE and the competing methods with the PPL metric. As shown in \cref{tab:GovReport}, on the tested Proof-pile and GovReport datasets, our HARPE achieves comparable or even better results compared to the state-of-the-art multi-stage methods and various single-stage methods.  This indicates that the proposed HARPE has the capability to handle long text sequences.
        
        % PPL results in \cref{tab:GovReport} indicate that HARPE achieved the lowest perplexity score on the Proof-pile and GovReport datasets, demonstrating exceptional performance. Methods based on ABF strategies also performed well in terms of perplexity, with relatively small performance differences among various ABF approaches. The advantage of HARPE is that it assigns training at different stages to different attention heads at the same time. This design enables the model to gradually adapt to long texts and ultimately reduce PPL.
        
        % HARPE's strength lies in its careful design of the base within the RoPE for each attention head. This design allows the positional information across different attention heads to complement one another, enabling the model to adaptively process information from various positions, ultimately leading to reduced PPL.
        
        %We use the upgraded Needle-in-a-Haystack test to evaluate the long-context performance of the methods listed in \cref{sub:baseline_systems}. As shown in \cref{tab:niah_methods_compare}, our HARPE model outperforms the LLaMA2 backbone by 6 points at 4k length. Additionally, it leads across all lengths and achieves an average score 5.5 points higher than the second-place ABF Muti-Stages method. HARPE continues to perform well on longer text evaluations. However, compared to YARN Self-Extend, its scores decline significantly beyond 32k in length. In particular, PI scores drop to zero starting from 32k. 

        Furthermore, we employ the upgraded Needle-in-a-Haystack test, as defined in the RULER benchmark \cite{hsieh2024ruler}, to evaluate the long-context relationship capturing performance of HARPE and its competitors. As shown in \cref{tab:niah_methods_compare}, HARPE significantly outperforms all listed methods. Notably, HARPE proves more effective than multi-stage approaches, surpassing the multi-stage ABF by 5.5\%. While typical single-stage methods, such as YARN and PI, fail as the context length increases, HARPE successfully extends the effective context length to 128K tokens. More details on the NiaH results, including scores for each of the 8 NiaH tasks (e.g., multi-key and multi-value), are provided in the Appendix \cref{tab:detail_niah_methods}. Simultaneously, we evaluate traditional NiaH tasks at a finer granularity, following the code in \cite{lwmgithub}. As shown in \cref{passkey_retrieval}, HARPE achieves a 100\% accuracy rate across various lengths within 128k tokens.
        
        % Detial results of all the NiaH tasks are summarized in \cref{tab:niah_methods_compare}.

        We also evaluate HARPE on the short-context benchmarks. Results in \cref{tab:benchmark_methods_compare} show that HARPE also yields comparable or even slightly better performance than competing methods in terms of average accuracy across 5 short-context tasks.

        %We select the top three methods: HARPE, ABF Single-Stage and three-stage ABF from  \cref{tab:niah_methods_compare}, for testing on short-context benchmark. The results in \cref{tab:benchmark_methods_compare} show that HARPE aslo surpasses the two ABF methods in average scores across the five short-context datasets.
        
    \subsection{Study of Various Base Schemes}\label{subsec:52_study}
        In this section, we evaluate the performance of two base selection methods for the head-specific RoPE bases in HARPE: uniform distribution and peak-valley opposition. For the uniform distribution method, we conduct two experiments with uniform \textbf{ascending and descending intervals} to analyze the impact of different base orders on model performance. For the peak-valley opposition method, as described in \cref{The searching algorithm of B_s} and \cref{eq:bc}, we test five variations with different base \textbf{strides (10k, 20k, 30k, 40k, 50k)} to further explore their effects.

        %In \cref{tab:niah_ablation} shows that HARPE evaluation results with different parameter settings consistently outperform the original LLaMA2 model, with a relatively stable overall trend. Comparing the two methods, the peak-valley opposition approach with $stride=30k$ surpasses the third-place model with uniform distribution by an average of 1.5 points in ascending order and 2.52 points in descending order in the upgraded NiaH test.

        The results of various HARPE configurations, along with the original LLaMA2 model, on the upgraded Needle-in-a-Haystack test are presented in \cref{tab:niah_ablation}. Under different RoPE base settings, our HARPE consistently outperforms the original LLaMA2-7B-Base model. Among the two methods evaluated, the peak-valley opposition approach with $stride=30k$ demonstrates the best performance, surpassing the next closest competitor by 1.25\%. As a result, we adopt the peak-valley approach with a stride of 30k for HARPE.

    \subsection{Comparative Results on RULER Evaluation}\label{subsec:PT}

    In this section, we evaluate HARPE against various open-source pre-trained models on a range of long-context tasks using the RULER benchmark. RULER is a comprehensive and widely recognized standard for long-context evaluation, comprising 13 tasks that include "needle in a haystack" as well as additional tasks such as Variable Tracing, Aggregation Ability, and Question Answering.
    
    As shown in \cref{tab:opensource_compare}, our comparison in 10 base models primarily involves 7B models, along with model using other architecture such as Jamba. HARPE surpasses all LLaMA2-based models and ranks 1st overall, surpassing the 2nd by 2.55\%. Notably, HARPE demonstrates a significant advantage in shorter context performance compared to the multi-stage ABF-trained LWM with a 1M fine-tuning length. Furthermore, HARPE consistently outperforms the YaRN model with a 128k fine-tuning length, achieving an average improvement of nearly 20 points across various lengths. Additionally, when compared to the llama-2-7b-80k model, which has the same training parameters and dataset but a shorter fine-tuning length of 80k, HARPE still shows superior performance in shorter context tasks with lengths less than 32k.

\section{Conclusion}  

In this paper, we present a novel single-stage continual pretraining method, HARPE, to enhance the long-context modeling capabilities of LLMs.
Specifically, our HARPE distributes the different training stages across different attention heads, and assigns different base values in the RoPE for different attention heads during continual pretraining stage.
Experimental results across 4 benchmarks demonstrate that HARPE outperforming or matching existing multi-stage methods in long-context modeling tasks, while maintaining comparable performance on short-context tasks.
In practical applications, our HARPE breaks the stage barrier, offering a simplified pipeline with minimal manual tuning and expertise, thereby streamlining the process of equipping LLMs with long-context capabilities.

\section{Limitations}
Despite that HARPE demonstrates promising results on benchmarks with long context lengths, limitation still remains. 
% First, due to resource constraints, our method was only evaluated with a context length of 128K and not tested with longer contexts such as 512K or 1M. 
Our research is primarily concentrated on the continual pretraining stage, leaving its applicability to other stages, such as supervised fine-tuning, unexplored. 
We will address those limitations in our future research.

% % Bibliography entries for the entire Anthology, followed by custom entries
\bibliography{custom}

\appendix

% \clearpage
% \section{Computational Overhead}

% HARPE does not increase inference costs, and its training cost is comparable to multi-stage training. When training LLaMa2 on 6B tokens using an 8×H800 machine, the multi-stage method takes 19H, while HARPE takes 20H. However, HARPE significantly reduces manual tuning, largely simplifying the process.

\section{Detail Scores}

\begin{table*}[t]
  \centering
  \resizebox{\textwidth}{!}{
  \begin{tabular}{ccccccccc}
    \toprule
    \centering
    \textbf{Method} &\textbf{Task} & \textbf{4k} & \textbf{8k} & \textbf{16k} & \textbf{32k} & \textbf{64k} & \textbf{128k} & \textbf{Avg.} \\
    \toprule
    \multirow{8}{*}{\textbf{PI}} & $niah\_single_1$ & 94.00 	 & 	35.00 	 & 	20.00 	 & 	0.00 	 & 	0.00 	 & 	0.00 	 & 	24.83 \\
    ~ & $niah\_single_2$ & 99.00 	 & 	46.00 	 & 	21.00 	 & 	0.00 	 & 	0.00 	 & 	0.00 	 & 	27.67 \\
    ~ & $niah\_single_3$ & 99.00 	 & 	45.00 	 & 	29.00 	 & 	0.00 	 & 	0.00 	 & 	0.00 	 & 	28.83 \\
    ~ & $niah\_multikey_1$ & 88.00 	 & 	39.00 	 & 	26.00 	 & 	0.00 	 & 	0.00 	 & 	0.00 	 & 	25.50 \\
    ~ & $niah\_multikey_2$ & 85.00 	 & 	14.00 	 & 	5.00 	 & 	0.00 	 & 	0.00 	 & 	0.00 	 & 	17.33 \\
    ~ & $niah\_multikey_3$ & 55.00 	 & 	2.00 	 & 	0.00 	 & 	0.00 	 & 	0.00 	 & 	0.00 	 & 	9.50  \\
    ~ & $niah\_multivalue$ & 32.25 	 & 	16.25 	 & 	17.00 	 & 	0.00 	 & 	0.00 	 & 	0.00 	 & 	10.92   \\
    ~ & $niah\_multiquery$ & 68.25 	 & 	15.50 	 & 	14.00 	 & 	0.00 	 & 	0.00 	 & 	0.00 	 & 	16.29  \\
    \midrule
    \multirow{8}{*}{\textbf{ABF Single-Stage}} & $niah\_single_1$ & 100.00 	 & 	100.00 	 & 	100.00 	 & 	99.00 	 & 	89.00 	 & 	91.00 	 & 	96.50  \\
    ~ & $niah\_single_2$ & 100.00 	 & 	100.00 	 & 	100.00 	 & 	100.00 	 & 	99.00 	 & 	93.00 	 & 	98.67 \\
    ~ & $niah\_single_3$ & 100.00 	 & 	100.00 	 & 	100.00 	 & 	100.00 	 & 	100.00 	 & 	88.00 	 & 	98.00 \\
    ~ & $niah\_multikey_1$ & 92.00 	 & 	93.00 	 & 	89.00 	 & 	91.00 	 & 	86.00 	 & 	86.00 	 & 	89.50 \\
    ~ & $niah\_multikey_2$ & 95.00 	 & 	98.00 	 & 	94.00 	 & 	82.00 	 & 	70.00 	 & 	37.00 	 & 	79.33 \\
    ~ & $niah\_multikey_3$ & 63.00 	 & 	54.00 	 & 	31.00 	 & 	16.00 	 & 	6.00 	 & 	3.00 	 & 	28.83 \\
    ~ & $niah\_multivalue$ & 92.00 	 & 	72.25 	 & 	66.00 	 & 	51.50 	 & 	34.75 	 & 	44.00 	 & 	60.08 \\
    ~ & $niah\_multiquery$ & 97.50 	 & 	93.00 	 & 	93.25 	 & 	84.75 	 & 	81.75 	 & 	59.75 	 & 	85.00 \\
    \midrule
    \multirow{8}{*}{\textbf{ABF Multi-Stage}} & $niah\_single_1$ & 100.00 	 & 	100.00 	 & 	100.00 	 & 	100.00 	 & 	96.00 	 & 	99.00 	 & 	99.17  \\
    ~ & $niah\_single_2$ & 100.00 	 & 	100.00 	 & 	100.00 	 & 	100.00 	 & 	100.00 	 & 	95.00 	 & 	99.17  \\
    ~ & $niah\_single_3$ & 100.00 	 & 	100.00 	 & 	100.00 	 & 	100.00 	 & 	100.00 	 & 	94.00 	 & 	99.00  \\
    ~ & $niah\_multikey_1$ & 95.00 	 & 	95.00 	 & 	95.00 	 & 	94.00 	 & 	83.00 	 & 	89.00 	 & 	91.83  \\
    ~ & $niah\_multikey_2$ & 96.00 	 & 	96.00 	 & 	91.00 	 & 	76.00 	 & 	73.00 	 & 	14.00 	 & 	74.33 \\
    ~ & $niah\_multikey_3$ & 79.00 	 & 	69.00 	 & 	48.00 	 & 	15.00 	 & 	8.00 	 & 	1.00 	 & 	36.67  \\
    ~ & $niah\_multivalue$ & 95.25 	 & 	81.00 	 & 	74.75 	 & 	55.50 	 & 	40.25 	 & 	40.75 	 & 	64.58   \\
    ~ & $niah\_multiquery$ & 96.25 	 & 	92.75 	 & 	91.50 	 & 	90.25 	 & 	82.00 	 & 	64.25 	 & 	86.17 \\
    \midrule
    \multirow{8}{*}{\textbf{YaRN}} & $niah\_single_1$ & 100.00 	 & 	100.00 	 & 	100.00 	 & 	100.00 	 & 	95.00 	 & 	0.00 	 & 	82.50  \\
    ~ & $niah\_single_2$ & 100.00 	 & 	100.00 	 & 	100.00 	 & 	87.00 	 & 	0.00 	 & 	0.00 	 & 	64.50  \\
    ~ & $niah\_single_3$ & 100.00 	 & 	100.00 	 & 	97.00 	 & 	65.00 	 & 	0.00 	 & 	0.00 	 & 	60.33  \\
    ~ & $niah\_multikey_1$ & 79.00 	 & 	69.00 	 & 	51.00 	 & 	30.00 	 & 	1.00 	 & 	0.00 	 & 	38.33  \\
    ~ & $niah\_multikey_2$ & 81.00 	 & 	64.00 	 & 	42.00 	 & 	12.00 	 & 	0.00 	 & 	0.00 	 & 	33.17 \\
    ~ & $niah\_multikey_3$ & 79.00 	 & 	69.00 	 & 	48.00 	 & 	15.00 	 & 	8.00 	 & 	1.00 	 & 	36.67  \\
    ~ & $niah\_multivalue$ & 86.00 	 & 	75.00 	 & 	59.50 	 & 	34.50 	 & 	3.50 	 & 	0.00 	 & 	43.08 \\
    ~ & $niah\_multiquery$ & 84.00 	 & 	72.25 	 & 	67.25 	 & 	42.75 	 & 	2.00 	 & 	0.00 	 & 	44.71  \\
    \midrule
    \multirow{8}{*}{\textbf{Self-Extend}} & $niah\_single_1$ & 100.00 	 & 	100.00 	 & 	100.00 	 & 	94.00 	 & 	11.00 	 & 	0.00 	 & 	67.50  \\
    ~ & $niah\_single_2$ & 100.00 	 & 	100.00 	 & 	96.00 	 & 	87.00 	 & 	0.00 	 & 	0.00 	 & 	63.83  \\
    ~ & $niah\_single_3$ & 100.00 	 & 	99.00 	 & 	90.00 	 & 	92.00 	 & 	0.00 	 & 	0.00 	 & 	63.50 \\
    ~ & $niah\_multikey_1$ & 78.00 	 & 	71.00 	 & 	56.00 	 & 	43.00 	 & 	0.00 	 & 	0.00 	 & 	41.33  \\
    ~ & $niah\_multikey_2$ & 28.00 	 & 	9.00 	 & 	3.00 	 & 	2.00 	 & 	0.00 	 & 	0.00 	 & 	7.00  \\
    ~ & $niah\_multikey_3$ & 26.00 	 & 	5.00 	 & 	3.00 	 & 	1.00 	 & 	0.00 	 & 	0.00 	 & 	5.83  \\
    ~ & $niah\_multivalue$ & 89.25 	 & 	64.50 	 & 	52.75 	 & 	46.25 	 & 	0.00 	 & 	0.00 	 & 	42.13  \\
    ~ & $niah\_multiquery$ & 90.50 	 & 	81.50 	 & 	70.00 	 & 	52.00 	 & 	0.00 	 & 	0.00 	 & 	49.00  \\
    \midrule
    \multirow{8}{*}{\textbf{HARPE}} & $niah\_single_1$ & 100.00 	 & 	100.00 	 & 	100.00 	 & 	100.00 	 & 	100.00 	 & 	100.00 	 & 	100.00   \\
    ~ & $niah\_single_2$ & 100.00 	 & 	100.00 	 & 	100.00 	 & 	100.00 	 & 	100.00 	 & 	100.00 	 & 	100.00  \\
    ~ & $niah\_single_3$ & 100.00 	 & 	100.00 	 & 	100.00 	 & 	100.00 	 & 	100.00 	 & 	99.00 	 & 	99.83 \\
    ~ & $niah\_multikey_1$ & 96.00 	 & 	97.00 	 & 	91.00 	 & 	92.00 	 & 	93.00 	 & 	90.00 	 & 	93.17  \\
    ~ & $niah\_multikey_2$ & 93.00 	 & 	96.00 	 & 	96.00 	 & 	87.00 	 & 	78.00 	 & 	46.00 	 & 	82.67  \\
    ~ & $niah\_multikey_3$ & 91.00 	 & 	91.00 	 & 	81.00 	 & 	38.00 	 & 	23.00 	 & 	6.00 	 & 	55.00  \\
    ~ & $niah\_multivalue$ & 98.50 	 & 	95.00 	 & 	92.25 	 & 	87.75 	 & 	61.50 	 & 	43.00 	 & 	79.67  \\
    ~ & $niah\_multiquery$ & 97.75 	 & 	96.00 	 & 	89.50 	 & 	88.50 	 & 	79.75 	 & 	53.50 	 & 	84.17  \\
    \bottomrule
  \end{tabular}
  }
  \caption{
    Detail Scores of the 8 upgraded Needle-in-a-Haystack Tasks Across Different Lengths. 
    }
  \label{tab:detail_niah_methods}
\end{table*}

%This is an appendix.

\end{document}